# PARALLEL SOFTWARE IMPLEMENTATION OF RECURSIVE MULTIDIMENSIONAL DIGITAL FILTERS FOR POINT-TARGET DETECTION IN CLUTTERED INFRARED SCENES

*Hugh L. Kennedy*

Defence and Systems Institute, University of South Australia
`hugh.kennedy@unisa.edu.au`

## ABSTRACT

A technique for the enhancement of point targets in clutter is described. The local 3-D spectrum at each pixel is estimated recursively. An optical flow-field for the textured background is then generated using the 3-D autocorrelation function and the local velocity estimates are used to apply high-pass velocity-selective spatiotemporal filters, with finite impulse responses (FIRs), to subtract the background clutter signal, leaving the foreground target signal, plus noise. Parallel software implementations using a multicore central processing unit (CPU) and a graphical processing unit (GPU) are investigated.

## 1. INTRODUCTION

Multidimensional filters provide a convenient framework for the extraction of tactical information from imaging sensor streams. Appropriately designed spatiotemporal *digital* filters, like *biological* vision systems, allow camouflaged objects to be extracted from cluttered scenes using foreground/background motion differences. Banks of velocity-tuned spatiotemporal filters are usually applied *globally* to sequences of image frames to separate (i.e. segment and classify) image regions according to their modes of motion, or lack thereof [1]-[5].

In this paper, high-pass velocity-tuned prediction-error filters (PEFs) are instead applied *locally* to each pixel in an attempt to suppress (or 'whiten') the background clutter and enhance point-like targets in the foreground (e.g. distant aircraft). High-order optimal filters offer diminishing returns when processing image data because the input signal is far from wide-sense stationary (in space and time), due to occlusion and discontinuities at object boundaries (i.e. edges). Therefore, a simple 3-D frequency-sampling method is adopted here, which avoids the need for the discretization of analog prototypes [5],[6]. Analytical expressions for the filter coefficients and frequency response are provided.

Each filter is tuned using the estimated (apparent) velocity of the background scene which is assumed to be dominated by heterogeneous textures, with predominantly low spatial-frequency content (e.g. distant cloud or terrain), possibly exhibiting *non-uniform* motion. The latter feature precludes the use of simple image registration-type techniques. Any number of optical-flow estimation techniques could have been employed for this purpose, such as gradient-based [7], phase-based [8], or block-matching methods (e.g. using the 2-D *cross*-correlation function) [9]; however, an alternative approach (using the local 3-D *auto*-correlation function) is instead used here [10]. This method was found to reduce the impact of noise and random variations, through the consideration of *multiple* 'stacked' frames, rather than simply a *pair* of consecutive frames.

A frequency-domain approach significantly reduces computational complexity by allowing the velocity-estimation and velocity-filtration stages to operate on a common spectrum, which is of course separable in all three dimensions. In previously reported approaches, a 2-D discrete Fourier transform (DFT) is applied to each new (spatial) frame using the fast Fourier transform (FFT) as a *batch* operation, then a (temporal) linear difference equation (LDE) is used to filter the 2-D spectrum to create a 3-D spatiotemporal (DFT/LDE) filter [11].

A somewhat similar approach is adopted here; however to better accommodate local variation, a 3-D (spatiotemporal) spectrum is estimated *recursively* for each pixel using the surrounding data within a 3-D analysis window. The sliding DFT (SDFT) is used in the spatial dimensions, which are of finite extent; whereas a deadbeat observer is used in the temporal dimension, which is effectively of infinite extent [12]. Both methods are implemented using filter banks with finite impulse responses (FIRs).

The SDFT was used for the spatial dimensions because it has a simple implementation – using a comb-filter cascaded with a bank of resonators [13]. Accumulated rounding errors due to pole-zero cancellation on the unit circle are negligible over the dimensions of a single frame [12].

The deadbeat observer implementation is slightly more complex and memory intensive as it uses a feedback loop with integrators in the forward path to drive steady-state errors to zero, which ensures that rounding errors do not accumulate over long periods of operation [12],[14].

Other approaches to the problem of point-target detection in cluttered infrared scenes tend to either focus on 1-D temporal filters [15], or 2-D spatial filters [16],[17]. When 3-D approaches *are* considered, different logic is typically applied in the spatial and temporal dimensions [18]. Use of multidimensional filters provides a clear and consistent framework for the joint consideration of both time and space [6].

The algorithm described in this paper is a preprocessor to 'whiten' sensor data that contain highly correlated data. A threshold is then applied to extract candidate target detections which may then be processed using a Bayesian tracker, to maintain continuity of target identity, without an excessively high false-track rate [19]. The velocity estimate at each pixel may also be used to assist the track initiation and data association functions of 'downstream' tracking processes. Alternatively, track-before-detect techniques could be applied [20]. These methods are known to provide very impressive SNR improvements for dim targets in noise; however, they also require the noise to be white and uncorrelated for satisfactory performance.

As the proposed preprocessor does not identify candidate target measurements, measurement-to-track data association is avoided; the processing sequence is therefore independent of the input data (i.e. the target and clutter density) resulting in a uniform computational load. Furthermore, the output at each pixel does not influence the output at nearby pixels, therefore parallel implementations are feasible. Reasonable rates of data throughput were achieved by parallelizing the (C++) software implementation for execution on multiple cores of the central processing unit (CPU) of a personal computer. Acceleration using graphics processing units (GPUs) was also investigated.

## 2. FORMULATION

The (monochrome) intensity $I$, of the background is estimated (or 'synthesized') at $\boldsymbol{n} - \acute{\boldsymbol{m}}$ in the sample domain via direct convolution using

$$\hat{I}(\boldsymbol{n} - \acute{\boldsymbol{m}}) = \sum_{m=0}^{M-1} H(\boldsymbol{m}; \acute{\boldsymbol{m}}, \boldsymbol{v}) I(\boldsymbol{n} - \boldsymbol{m}) \quad (1)$$

where $H(\boldsymbol{m}; \acute{\boldsymbol{m}}, \boldsymbol{v})$ are the coefficients of a finite-impulse-response (FIR) filter, tuned to the velocity $\boldsymbol{v}$, with a nominal group 'delay' in each dimension of $\acute{\boldsymbol{m}} = [\acute{m}_x, \acute{m}_y, \acute{m}_z]$ samples. The image is indexed using $\boldsymbol{n} = [n_x, n_y, n_z]$ for $n = 0 \dots N - 1$ in the spatial $(x, y)$ and temporal dimensions $(z)$. The finite 'analysis' window is indexed in the opposite sense, as per the usual convention, using $\boldsymbol{m} = [m_x, m_y, m_z]$ for $m = 0 \dots M - 1$ in each dimension where $M$ is odd, i.e. $M = 2K + 1$. Thus the origin of the analysis window at $\boldsymbol{m} = [0,0,0]$ is at $\boldsymbol{n} = [n_x, n_y, n_z]$. The estimated background intensity $\hat{I}$, is then subtracted from the actual intensity $I$, at $\boldsymbol{n} - \acute{\boldsymbol{m}}$ to (ideally) yield foreground features plus noise, if the assumed model used to design the filter holds [10].

The background is assumed to be a band-limited signal in the spatial dimensions with normalized frequency (cycles/sample) of $|f| \leq B/M$ where $B < K$. If only discrete and uniformly-spaced frequencies are considered i.e. for $k = -B \dots + B$, then

$$H(\boldsymbol{m}; \acute{\boldsymbol{m}}, \boldsymbol{v}) = \frac{W_x W_y}{M_x M_y M_z} \times$$
$$\mathcal{D}_{W_x}([m_x - \acute{m}_x - v_x(m_z - \acute{m}_z)]/M_x) \times$$
$$\mathcal{D}_{W_y}([m_y - \acute{m}_y - v_y(m_z - \acute{m}_z)]/M_y) \quad (2)$$

where $\mathcal{D}_W$ is the Dirichlet kernel of order $W = 2B + 1$, defined here as

$$\mathcal{D}_W(w) = \frac{\sin(\pi W w)}{W \sin(\pi w)}. \quad (3)$$

Computational efficiency is increased for band-limited signals, if analysis and filtering are performed in the frequency domain using

$$\hat{I}(\boldsymbol{n} - \acute{\boldsymbol{m}}) = \sum_{k=-B}^{+B} \mathcal{H}(\boldsymbol{k}; \acute{\boldsymbol{m}}, \boldsymbol{v}) S(\boldsymbol{k}; \boldsymbol{n}) \quad (4)$$

for $k_{xy} = -B_{xy} \dots + B_{xy}$ and $k_z = -K_z \dots + K_z$ and where $\mathcal{H}(\boldsymbol{k}; \acute{\boldsymbol{m}}, \boldsymbol{v})$ and $S(\boldsymbol{k}; \boldsymbol{n})$ are frequency-domain representations of $H$ and a windowed block of $I$, which are both transformed via a DFT using

$$\mathcal{H}(\boldsymbol{k}; \acute{\boldsymbol{m}}, \boldsymbol{v}) = \sum_{m=0}^{M-1} F^*(\boldsymbol{m}; \boldsymbol{k}) H(\boldsymbol{m}; \acute{\boldsymbol{m}}, \boldsymbol{v}) \quad (5)$$

and

$$S(\boldsymbol{k}; \boldsymbol{n}) = \sum_{m=0}^{M-1} F(\boldsymbol{m}; \boldsymbol{k}) I(\boldsymbol{n} - \boldsymbol{m}) \quad (6)$$

with

$$F(\boldsymbol{m}; \boldsymbol{k}) = \frac{1}{\sqrt{M_x M_y M_z}} e^{j2\pi\left(\frac{k_x}{M_x}m_x + \frac{k_y}{M_y}m_y + \frac{k_z}{M_z}m_z\right)}. \quad (7)$$

Note that the asterisk superscript denotes complex conjugation. The frequency response of the filter is found by summing over each of the frequency components comprising the background

$$\mathcal{Q}(\boldsymbol{f}; \acute{\boldsymbol{m}}, \boldsymbol{v}) = \frac{1}{\sqrt{M_x M_y M_z}} \sum_{k_y=-B_y}^{+B_y} \sum_{k_x=-B_x}^{+B_x} q(\boldsymbol{f}) \quad (8a)$$

where

$$q(\boldsymbol{f}) = b_x b_y b_z \, c_x(f_x) c_y(f_y) c_z(f_z) d_x(f_x) d_y(f_y) d_z(f_z) \quad (8b)$$

with

$$b_x = e^{-j2\pi \frac{k_x}{M_x}(\acute{m}_x - \Delta_x)} \quad (8c)$$
$$b_y = e^{-j2\pi \frac{k_y}{M_y}(\acute{m}_y - \Delta_y)} \quad (8d)$$
$$b_z = e^{+j2\pi(v_x k_x/M_x + v_y k_y/M_y)(\acute{m}_z - \Delta_z)} \quad (8e)$$
$$c_x(f_x) = e^{-j2\pi f_x \Delta_x} \quad (8f)$$
$$c_y(f_y) = e^{-j2\pi f_y \Delta_y} \quad (8g)$$
$$c_z(f_z) = e^{-j2\pi f_z \Delta_x} \quad (8h)$$
$$d_x(f_x) = \mathcal{D}_{M_x}(f_x - k_x/M_x) \quad (8i)$$
$$d_y(f_y) = \mathcal{D}_{M_y}(f_y - k_y/M_y) \quad (8j)$$
$$d_z(f_z) = \mathcal{D}_{M_z}(f_z + v_x k_x/M_x + v_y k_y/M_y) \quad (8k)$$

The frequency-domain filter coefficients $\mathcal{H}(\boldsymbol{k}; \acute{\boldsymbol{m}}, \boldsymbol{v})$, are found by evaluating $\mathcal{Q}(\boldsymbol{f}; \acute{\boldsymbol{m}}, \boldsymbol{v})$ at the each of the DFT bins used to transform $I$ in (6), i.e. by substituting $\boldsymbol{f} = [k_x/M_x, k_y/M_y, k_z/M_z]$ into (8).

The velocity-induced frequency-shift, that tilts the spatial frequencies out of the $xy$ plane (where $f_z = 0$) according to $f_z = -v_x f_x - v_y f_y$, results in a plane passing through frequencies in the $z$ dimension that do not necessarily coincide with the discrete frequency bins at $f_z = k_z/M_z$ of the DFT [4]. The Dirichlet kernel of order $M_z$ in the $z$ dimension is therefore required to capture the 'sidelobes' that result when the 'energy' of each sinusoidal $xy$ component 'spills' into adjacent bins due to the misalignment of the nodes of $\mathcal{D}_{M_z}(f_z)$ with the bins at $f_z = k_z/M_z$. In contrast, the nodes of $\mathcal{D}_{M_x}(f_x)$ and $\mathcal{D}_{M_y}(f_y)$ *do* coincide with the DFT bins at $f_x = k_x/M_x$ and $f_y = k_y/M_y$, therefore the Dirichlet kernel is only used to interpolate the filter response in between the DFT bins in the spatial dimensions. The $b$ factors in (8) perform the synthesis operation; the $c$ factors and the $\Delta_{xyz}$ constants compensate for the displacement of the sample-domain origin from the center of the analysis window – as 'displacement' in the sample domain is 'modulation' in the frequency domain. The modulation required uses $\Delta = (M-1)/2$ in each dimension.

Only the low-band frequencies comprising the background signal are used for filtering purposes; however, all 'measureable' frequencies are evaluated when the DFT is computed because high-frequency content is required to capture edges and other spatial details for reliable optical-flow determination. Similarly, near-DC frequencies contain little-to-no useful flow information, therefore spatial components with $k = 0$ are omitted.

The optical flow field is determined using the local estimate of the 3-D autocorrelation function. The local velocity is estimated using $\hat{v}_x = \hat{l}_x/l_z$ and $\hat{v}_y = \hat{l}_y/l_z$ where $\hat{l}_x$ and $\hat{l}_y$ are the displacements that maximize $R$ in the $l_z = 1$ 'slice'. Finer velocity resolution is achieved by evaluating $R$ at non-integer (interpolating) displacements. Using $l \ll M - 1$ is recommended to reduce the impact of the assumed cyclic boundary condition, as zero-padding is not used. To minimize the impact of point-targets in the foreground on the background velocity estimate, the spatial extent of the analysis window also needs to be much larger than the spatial extent of the target. Recursively smoothing $R$ at each pixel using a temporal first-order filter, with a real pole at $\alpha$ ($0 < \alpha < 1$), was also found to be beneficial in this respect.

The power density spectrum for the $\boldsymbol{k}$th frequency bin, for the pixel located at $\boldsymbol{n} - \acute{\boldsymbol{m}}$, is estimated using
$$P(\boldsymbol{k}; \boldsymbol{n}) = \acute{S}^*(\boldsymbol{k}; \boldsymbol{n})\acute{S}(\boldsymbol{k}; \boldsymbol{n}). \qquad (9)$$
A Hann window is applied to the high-pass filtered spectrum to yield $\acute{S}$. This operation is not required for filtering but it was found to significantly improve velocity estimation. This operation is applied as a convolution in the frequency domain. Even when the windows are separated in each dimension, this is unfortunately quite an 'expensive' process. The (high-pass filtered) 3-D autocorrelation function is then evaluated using
$$R(\boldsymbol{l}; \boldsymbol{n}) = \sqrt{M_x M_y M_z} \sum_{\boldsymbol{k}=-K}^{+K} F^*(\boldsymbol{l}, \boldsymbol{k}) P(\boldsymbol{k}; \boldsymbol{n}). \qquad (10)$$
The velocity estimate is derived from the smoothed version of $R$ (i.e. $\acute{R}$) where $\acute{R}_{n_z} = (1-\alpha)R + \alpha \acute{R}_{n_z-1}$.

## 3. PARAMETERIZATION

The following filter parameters were used to process the real and simulated data: $K_x = K_y = 4, K_z = 2$ (determines filter kernel dimensions); $B_x = B_y = 3$ (determines filter bandwidth); $\acute{m}_x = K_x, \acute{m}_y = K_y, \acute{m}_z = K_z$ (for a linear phase filter, but a two frame filter latency); $l_x = l_y = [-8, -7, \ldots + 7, +8]/4$ (for a maximum velocity of 2 pixels per frame and a velocity grid spacing of 1/4); $\alpha = e^{-1/10} \cong 0.9$ (for low-gain autocorrelation smoother); $N_x = N_y = 64$ (image dimensions).

## 5. IMPLEMENTATION

Various realizations of the proposed whitening filter were implemented on a personal computer with a 64 bit operating system and an Intel ® i7-4810MQ CPU @ 2.8GHz with 4 physical cores, each functioning as two logical processors, for a total of 8 concurrent threads. All realizations gave identical outputs (considering the numerical precision used) when processing pre-generated/pre-recorded data.

A MATALB ® algorithm prototype was coded using only the core MATALB (R2013b) engine (i.e. no toolboxes). With recursively-generated local spectra and 'vectorized' code, this instantiation achieved a processing rate of 2.3 frames per second (Hz). Monitoring processor utilization during execution revealed that the computational load was distributed uniformly over 8 concurrent threads.

Various C++ instantiations were then coded using Visual Studio 2012 ®. Single-precision floating-point variables were used in all cases. The baseline serial instantiation, with recursive DFT filters, achieved a processing rate of 6.3 Hz. Using `parallel_for` loops, provided by the Parallel Patterns Library (PPL), increased the processing rate to 43.7 Hz by distributing the load over multiple threads on the CPU. This result is very close to the maximum theoretical speedup of 8x. Parallelizing the recursive spectrum generation stage only resulted in a slight speedup. The significant speedup observed was mainly due to the parallel processing of each image row during the velocity estimation and background filtration stages, where each thread has much more work to do.

Using `parallel_for_each` loops, provided by the Accelerated Massive Parallelism (AMP) library, to enable GPU utilization, was somewhat less gratifying. Language constraints – such as: no complex types, the need to use `array_view` or `array` containers instead of pointers, a limit on the maximum number of arrays that can be used, etc. – meant that substantial re-coding was required. These constraints also placed an upper limit on the complexity of the code within the body of each loop, which reduced the total work that each thread was able to do. As a consequence, the impact of parallel overheads (e.g. data transfer and synchronization) was more noticeable. All attempts to parallelize various parts of the code (e.g. spectrum generation and/or window application) resulted in processing rates that were slower than the baseline serial implementation. The CPU-"integrated" GPU and the external "high-performance" GPU both gave similar results.

In this particular problem, a GPU implementation provides the greatest speedup when a naïve non-recursive approach to the DFT is adopted. Generation of a non-recursive local spectrum, for each pixel in *series*, proceeds at a rate of 0.25 Hz; whereas processing the pixels in *parallel* on the GPU progresses at a rate of 5.9 Hz – a speedup of nearly 24x. For comparison: recursive-DFT processing-rates for the serial CPU, parallel CPU, and parallel GPU implementations are 87.7 Hz, 186.6 Hz, and 18.6 Hz, respectively. Thus the recursive DFT 'beats' the parallel GPU 'to the punch'.

### 3. SIMULATION

A synthetic background, translating with a velocity of $v_{\text{clt}} = [1.625, 0.625]$ (pixels per frame), i.e. midway between velocity 'bins', was generated using 25 pseudo-randomly generated components. The frequency of each component in each spatial dimension was drawn from a uniform distribution over the interval $f = [-2, +2]/9$ (cycles per pixel), with random phase, and a magnitude of 0.1. A DC offset of 10.0 was added to the background. A foreground point target with a velocity of $v_{\text{tgt}} = [-0.625, -0.375]$ and a maximum intensity 1.0 of was then injected. A Gaussian point-spread function, with a standard deviation of 1 pixel was used to 'smear' the target signal over adjacent pixels. The target obscured the background (i.e. not additive) and the target was truncated when its intensity fell below 0.1. Its initial position was set so that it is located at the center of the field of view in the final frame. Zero-mean Gaussian noise, with a standard deviation of 0.1 was then added to the image.

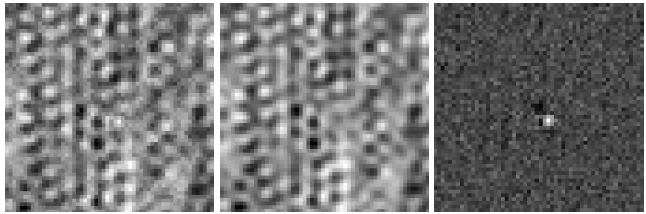

**Fig. 1.** Simulated data example. Left: raw input data, delayed by two frames so that it is 'aligned' with filter output. Middle: output of the background prediction filter. Right: output of prediction-error filter (i.e. the difference between the left and middle images), clearly showing the foreground target near the center of the frame.

### 4. APPLICATION

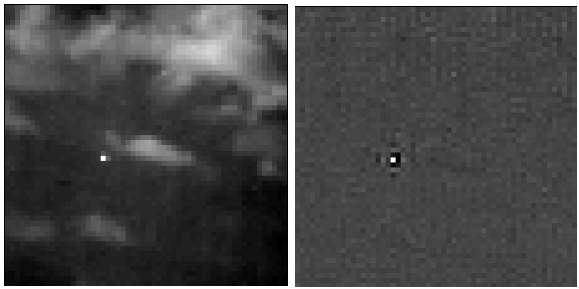

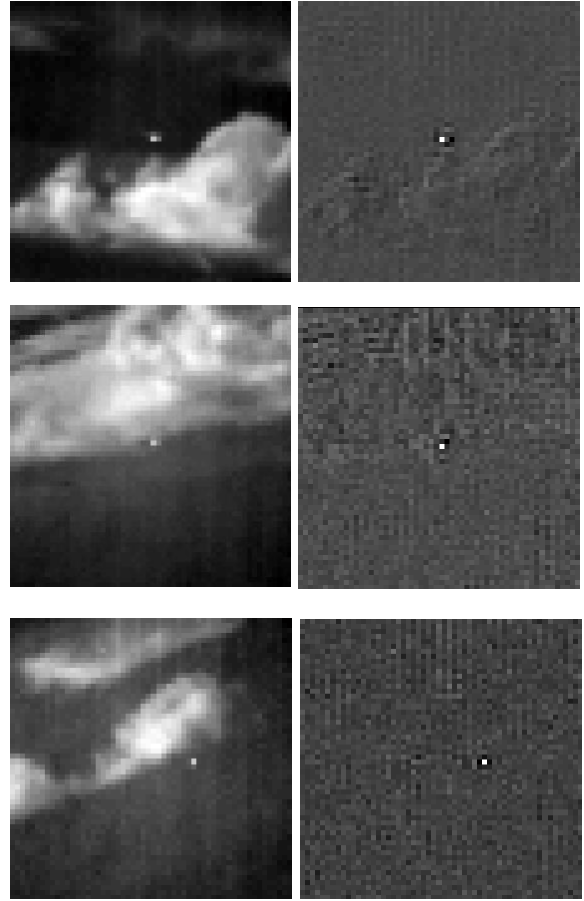

**Fig. 2.** Real infrared data example. Post-processed. Left column: raw input data (delayed by 2 frames); Right column: output of prediction-error filter. Top & bottom rows: stationary camera (background motionless); Middle rows: panning camera (target motionless).

### 6. CONCLUSION

Now that transistor densities in silicon chips are approaching practical limits and processor clock frequencies plateauing, engineers are increasingly relying on parallel processing to accelerate digital signal processing applications. Development in this area is currently progressing rapidly on two fronts: CPU and GPU. Both are the focus of considerable investment and manufacturers are competing for dominance. The most appropriate parallelization path depends on many factors such as the nature of the problem, the specifications of the hardware and the approach of the programmer [21]. For the multidimensional digital filter presented in this paper, CPU acceleration was found to be optimal, with a near-ideal speedup of 8x achieved with very little effort. The proposed filter was found to be very effective for enhancing point-targets, set against cluttered backgrounds, in infrared image sequences.